\documentclass[runningheads]{llncs}

\usepackage[T1]{fontenc}
\usepackage{graphicx}
\usepackage{amsmath,amssymb}
\usepackage{color}
\usepackage{multirow}
\usepackage{hyperref}

\makeatletter
\def\thanks#1{\protected@xdef\@thanks{\@thanks
        \protect\footnotetext{#1}}}
\makeatother

\begin{document}

\title{APAUNet: Axis Projection Attention UNet for Small Target in 3D Medical Segmentation}

\titlerunning{APAUNet: Axis Projection Attention UNet}

\author{
Yuncheng Jiang\inst{1,2,3,\star}
\and
Zixun Zhang\inst{1,2,3,\star}\thanks{$\star$~Equal contributions. Code is available at \href{https://github.com/zx33/APAUNet}{github.com/zx33/APAUNet}.}
\and
Shixi Qin\inst{1,2,3}
\and
Yao Guo\inst{4}
\and
\\
Zhen Li\inst{2,1,3,\dag}\thanks{\dag~Corresponding Author}
\and
Shuguang Cui\inst{2,1,5}
}

\authorrunning{Y. Jiang et al.}

\institute{
FNii, CUHK-Shenzhen, Guangdong, China \\
\email{\{yunchengjiang@link.,zixunzhang@link.,lizhen@\}cuhk.edu.cn}
\and
SSE, CUHK-Shenzhen, Guangdong, China
\and
SRIBD, CUHK-Shenzhen, Guangdong, China
\and
Shanghai Jiao Tong Univerisity, Shanghai, China
\and
Pengcheng Laboratory, Shenzhen,Guangdong, China
}

\maketitle

\begin{abstract}

In 3D medical image segmentation, small targets segmentation is crucial for diagnosis but still faces challenges.
In this paper, we propose the {\textbf{A}}xis {\textbf{P}}rojection {\textbf{A}}ttention UNet, named \textbf{APAUNet}, for 3D medical image segmentation, especially for small targets. 
Considering the large proportion of the background in the 3D feature space, we introduce a projection strategy to project the 3D features into three orthogonal 2D planes to capture the contextual attention from different views. 
In this way, we can filter out the redundant feature information and mitigate the loss of critical information for small lesions in 3D scans. 
Then we utilize a dimension hybridization strategy to fuse the 3D features with attention from different axes and merge them by a weighted summation to adaptively learn the importance of different perspectives. 
Finally, in the APA Decoder, we concatenate both high and low resolution features in the 2D projection process, thereby obtaining more precise multi-scale information, which is vital for small lesion segmentation. 
Quantitative and qualitative experimental results on two public datasets (BTCV and MSD) demonstrate that our proposed APAUNet outperforms the other methods. 
Concretely, our APAUNet achieves an average dice score of 87.84 on BTCV, 84.48 on MSD-Liver and 69.13 on MSD-Pancreas, and significantly surpass the previous SOTA methods on small targets.


\end{abstract}

\section{Introduction}

Medical image segmentation, which aims to automatically and accurately diagnose lesion and organ regions in either 2D or 3D medical images, is one of the critical steps for developing image-guided diagnostic and surgical systems.
In practice, compared to large targets, such as organs, small targets like tumors or polyps are more important for diagnosis, but also prone to be ignored. 
In this paper, we focus on 3D medical image segmentation (CT/MRI), with an emphasis on small lesions. 
This task is challenging mainly due to the following two aspects: 1) severe class imbalance of foreground (lesions) and background (entire 3D scans); 2) large variances in shape, location, and size of organs/lesions.

\begin{figure}[t]
	\centering
	\includegraphics[width=0.75\linewidth]{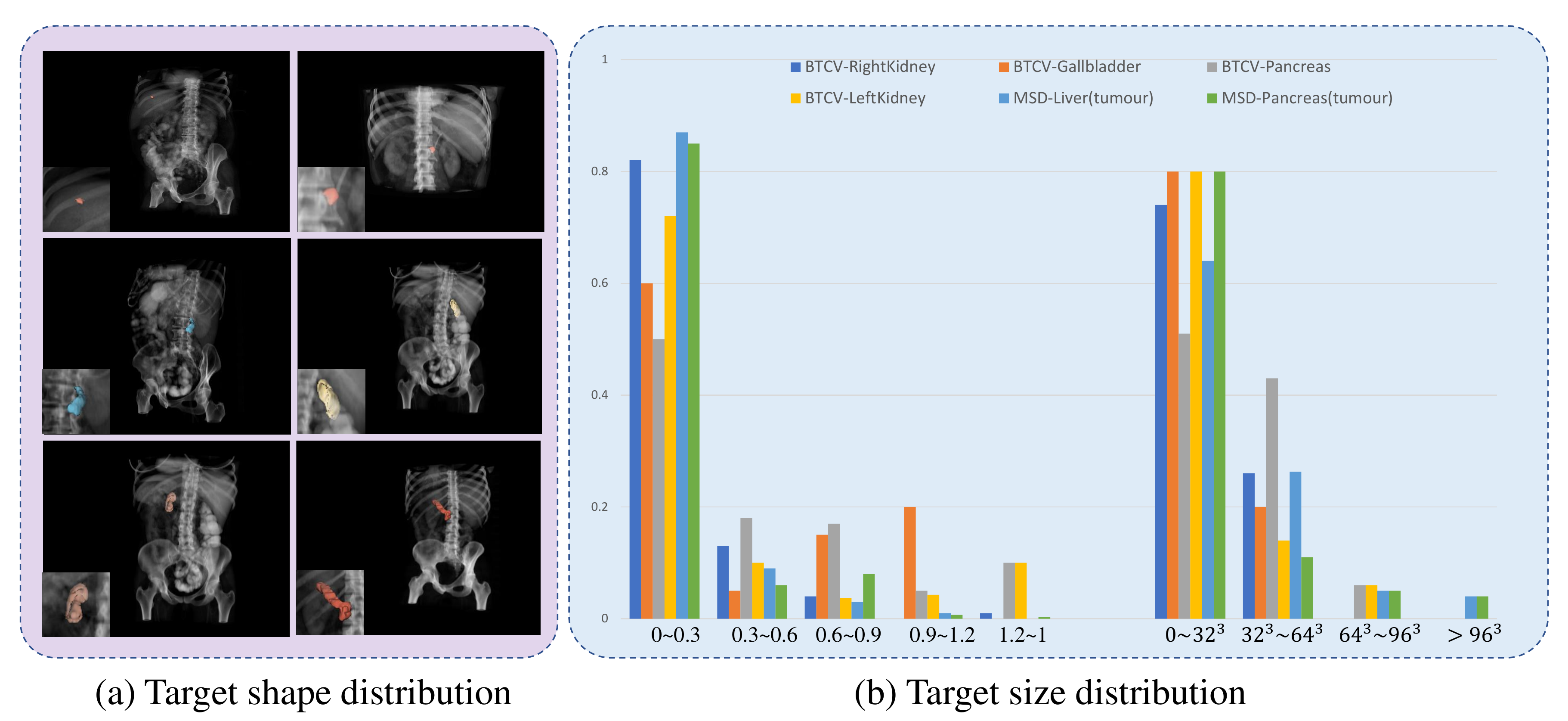}
	\caption{
	Target shape samples and size distribution of MSD and synapse multi-organ segmentation dataset. 
	(a) 6 example organs from synapse multi-organ segmentation dataset. 
	(b) The target size distribution. 
	The $x$-axis is the target size interval, and the $y$-axis is the proportion (\%) of corresponding samples to the whole dataset. 
	The left part shows the relative proportion (\%) of the target size to the whole input, 
	while the right part shows the absolute size of the target with a interval step of 32 voxels. 
	It can be observed that the relative target sizes of most samples in all the 6 categories are less than 0.6\% with various shapes.}
	\label{bar}
\end{figure}

Recent progress in medical image segmentation has mainly been based on UNet~\cite{ronneberger2015u}, which applies a U-shaped structure with skip-connections to merge multi-scale features. 
However, due to the inductive bias of the locality of the convolutions, the U-shaped networks still suffer from the limited representation ability. 
Some studies utilized a coarse-to-fine segmentation framework~\cite{zhu20183d}. These approaches refine the final segmentation in the fine stage, by shrinking input features to the region of interest (ROI) predicted in the coarse stage. 
Also, instead of using vanilla 3D convolutions, some works tried to explore a 2.5D fashions~\cite{wang201925D}, which performed 2D convolutions on $xy$-axis at the low-level layers of the network and 3D convolutions on the high-level layers. 
Other works attempted to use an ensemble of 2D and 3D strategies, which fuses the 2D predictions from different views with 3D predictions to get better results~\cite{zheng20192D3Dparallel} or refine the 2D predictions using 3D convolutions~\cite{xia20182D3Dserial}. 
Besides, inspired by the great success of Transformers, some works explored the feasibility of applying self-attention into medical images by integrating CNN-based architectures with Transformer-like modules~\cite{gao2021utnet,hatamizadeh2022unetr,CoTr2021} to capture the patch-level contextual information. 

Although previous methods have achieved remarkable progress, there are still some issues: 
1) The 2.5D or the ensemble of 2D and 3D methods still suffer from the limited representation ability since the 2D phases only extract features from two axes while ignoring the information from the other axis, which worsens the final segmentation prediction. 
Also, the two-stage designs are difficult for end-to-end training and require more computational resources. 
2) Transformer-like models require higher computational cost on self-attention, and thus, have limited applications in 3D scenarios. Moreover, these models only learn the attentive interactions between patches, yet ignoring the local pattern inside the patch. 
3) In addition, the imbalance between target and background has been ignored, which is vital for 3D medical segmentation. 
As shown in Fig.~\ref{bar}, on MSD chanllenge and BTCV datasets, the majority samples of the tumour target and small organ target are smaller than 0.6\% to the whole 3D scans with various shapes.

\begin{figure}[t]
	\centering
	\includegraphics[width=0.75\linewidth]{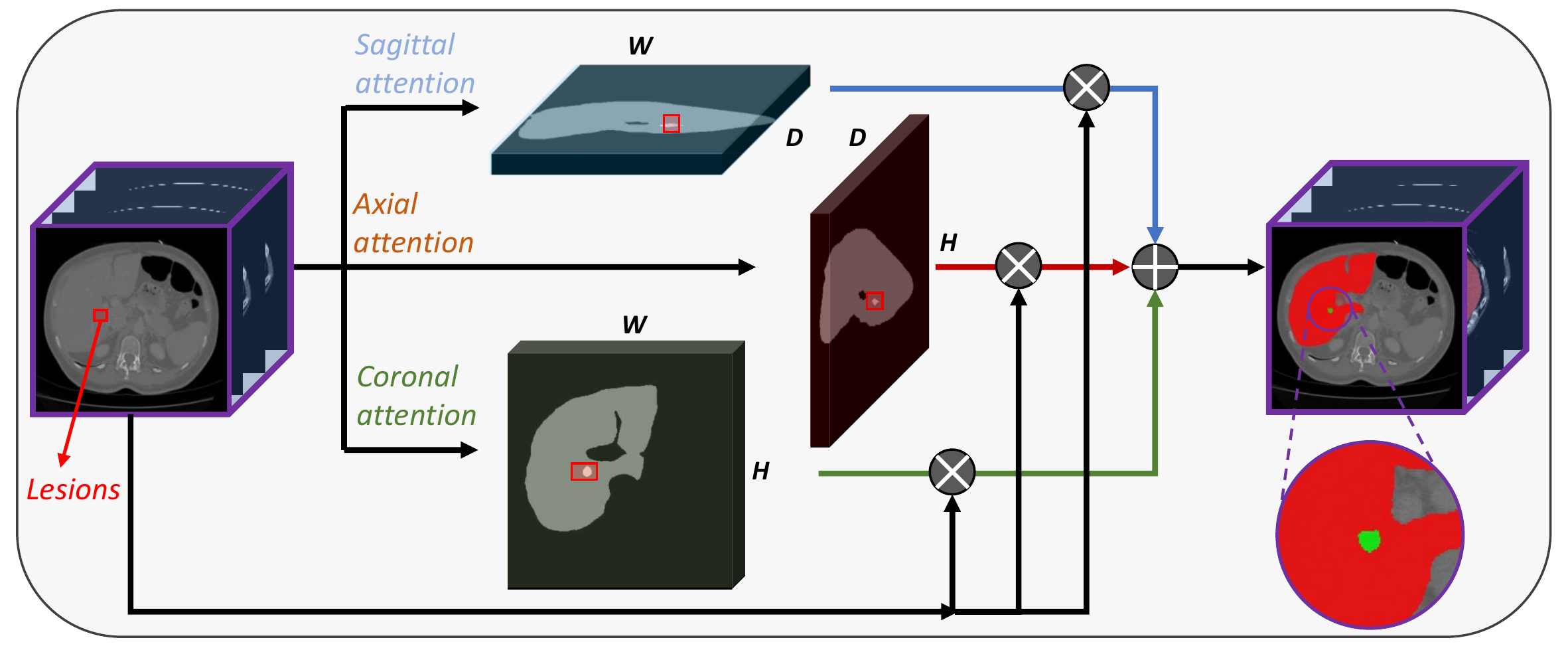}
	\caption{
	In our APAUNet, we first project the 3D features into three orthogonal 2D planes to capture local contextual attentions from the three 2D perspectives, and then fuse them with the original 3D features. 
	Finally, we adaptively fuse the features by weighted summation.
	}
	\label{teaser}
\end{figure}

In this paper, we propose an Axis Projection Attention (APA) UNet, named APAUNet, which utilizes an orthogonal projection strategy and a dimension hybridization strategy to overcome the aforementioned challenges.
Specifically, our APAUNet follows the established design of 3D-UNet but replaces the main and functional component, the 3D convolution based encoder/decoder layers, with our APA encoder/decoder modules. 
In the \textit{APA encoder}, the initial 3D feature maps are projected to three orthogonal 2D planes, i.e., \textit{sagittal}, \textit{axial}, and \textit{coronal} views. 
Such a projection operation could mitigate the loss of critical information for small lesions in 3D scans. 
For instance, the original foreground-background area ratio of 3D features is $O(1/n^3)$ before the projection, but after projection, the ratio can be promoted to $O(1/n^2)$. 
Afterwards, we extract the local contextual 2D attention along the projected features to perform the asymmetric feature extraction and fuse them with the original 3D features. 
Eventually, the fused features of three axes are summed as the final output by three learnable factors, as shown in Fig.~\ref{teaser}. 
Correspondingly, our \textit{APA decoder} follows the same philosophy as the APA encoder but takes input features from two resolution levels. In this way, the decoder can effectively leverage the contextual information of multi-scale features.
Furthermore, we also utilize an oversampling strategy to ensure the occurrence of foregrounds in each batch during the training process.

In summary, our contributions are in three-fold:
(1) We propose the Axis Projection Attention UNet. APAUNet utilities the orthogonal projection strategy to enhance the asymmetric projection attention and feature extraction. 
(2) We introduce a novel dimension hybridization strategy to fuse 2D and 3D attention maps for better contextual representation in both encoder and decoder blocks. Besides, we further leverage a multi-resolution fusion strategy into decoder blocks for context enhancement. 
(3) Extensive experiments on Synapse multi-organ segmentation (BTCV)~\cite{miccaibtcv} and Medical Segmentation Decathlon (MSD) challenge~\cite{simpson2019large} datasets demonstrate the effectiveness and efficiency of our APAUNet, especially on small targets.

\section{Related Work}

\subsection{CNN-based Medical Image Segmentation}

CNNs, serving as the standard model of medical image segmentation, have been extensively studied in the past.
The typical U-shaped network, U-Net~\cite{ronneberger2015u}, which consists of a symmetric encoder and decoder network with skip-connections, has become a common choice for medical image analysis.
Afterwards, different variations of U-Net were proposed, such as Res-UNet~\cite{8589312}, and Dense-UNet~\cite{li2018h}. 
Besides, there are also some studies using AutoML to search for UNet architectures or an ensemble of 2D and 3D features, 
e.g., C2FNAS~\cite{yu2020c2fnas} uses a two stages NAS to search for the 3D architecture, 
and \cite{zheng20192D3Dparallel} utilizes meta learner to learn the ensemble of 2D and 3D features. 
Although these architectures have achieved remarkable progress in various 2D and 3D medical image segmentation tasks, they lack the capability to learn the global context and long-range spatial dependencies, even though followed by down-sampling operations. Thus, it leads to the degraded performance on the challenging task of small lesion segmentation.

\subsection{Attention Mechanism for Medical Imaging}

Attention mechanisms have been widely applied to segmentation networks, which can be categorized into two branches.
The first branch is the hard attention, which typically uses a coarse-to-fine framework for segmentation tasks.
\cite{zhu20183d} exploited two parallel FCNs to first detect the ROI of input features, then conducted the fine-grained segmentation over these cropped ROI patches for volumetric medical image segmentation.
RA-UNet~\cite{10.3389/fbioe.2020.605132} introduced a residual attention module that adaptively combined multi-level features, which precisely extracted the liver region and then segmented tumours in this region.
However, these hard attention methods usually need extensive trainable parameters and can be difficult to converge, which are not efficient for 3D medical segmentation tasks.
The second branch is the adoption of the self-attention mechanism. 
One of the early attempts was the Attention U-Net~\cite{oktay2018attention}, which utilized an attention gate to suppress irrelevant regions of the feature map while highlighting salient features. 
UTNet~\cite{gao2021utnet} adopted efficient self-attention encoder and decoder to alleviate the computational cost for 2D medical image segmentation.
UNETR~\cite{hatamizadeh2022unetr} further employed a pure transformer by introduceding a multi-head self-attention mechanism into the 3D-UNet structure, taking advantage of both Transformers and CNNs to learn the sequential global features.
Nonetheless, considering the class imbalance issue of small lesions and large variants of organs, the methods mentioned above are not effective enough. 
To this end, our work aims at developing an efficient approach, thoroughly taking advantage of the attention mechanism for specifically small lesion segmentation in 3D scans.

\subsection{Small Target Segmentation}

Segmentation of small objects, with limited available features and imbalanced sample distribution, is a common challenge in computer vision tasks, especially for medical images. 
Many previous studies have explored the solutions for natural images. 
For example, improved data augmentation methods~\cite{zhang2018mixup,yun2019cutmix} were proposed to increase the diversity of data and enhance the model generalization ability. 
In addition, advanced feature fusion techniques were adopted to better capture the small objects~\cite{chen2017deeplab}.
However, few works investigate this issue for medical images, and the obtained performance is usually less superior. 
For instance, \cite{xu2020automatic} utilized a reinforcement learning (RL) based search method to construct the optimal augmentation policy for small lesions segmentation. 
UNet 3+~\cite{9053405} exploited the full-scale skip-connections to extract small-scale features efficiently. 
C2FNAS \cite{yu2020c2fnas} and DINTS \cite{he2021dints} used Neural Architecture Search (NAS) to find an appropriate network for small features extraction. 
MAD-UNet \cite{mad-unet} stacked multi-scale convolutions with attention modules to reduce the effects of intra-class inconsistency and enrich the contextual information. 
Nonetheless, these approaches are neither superior in performance nor computationally efficient, which motivates us to make an in-depth investigation of small object segmentation in 3D medical scans.

\section{Methodology}

\begin{figure}[t]
    \centering
    \includegraphics[width=0.8\linewidth]{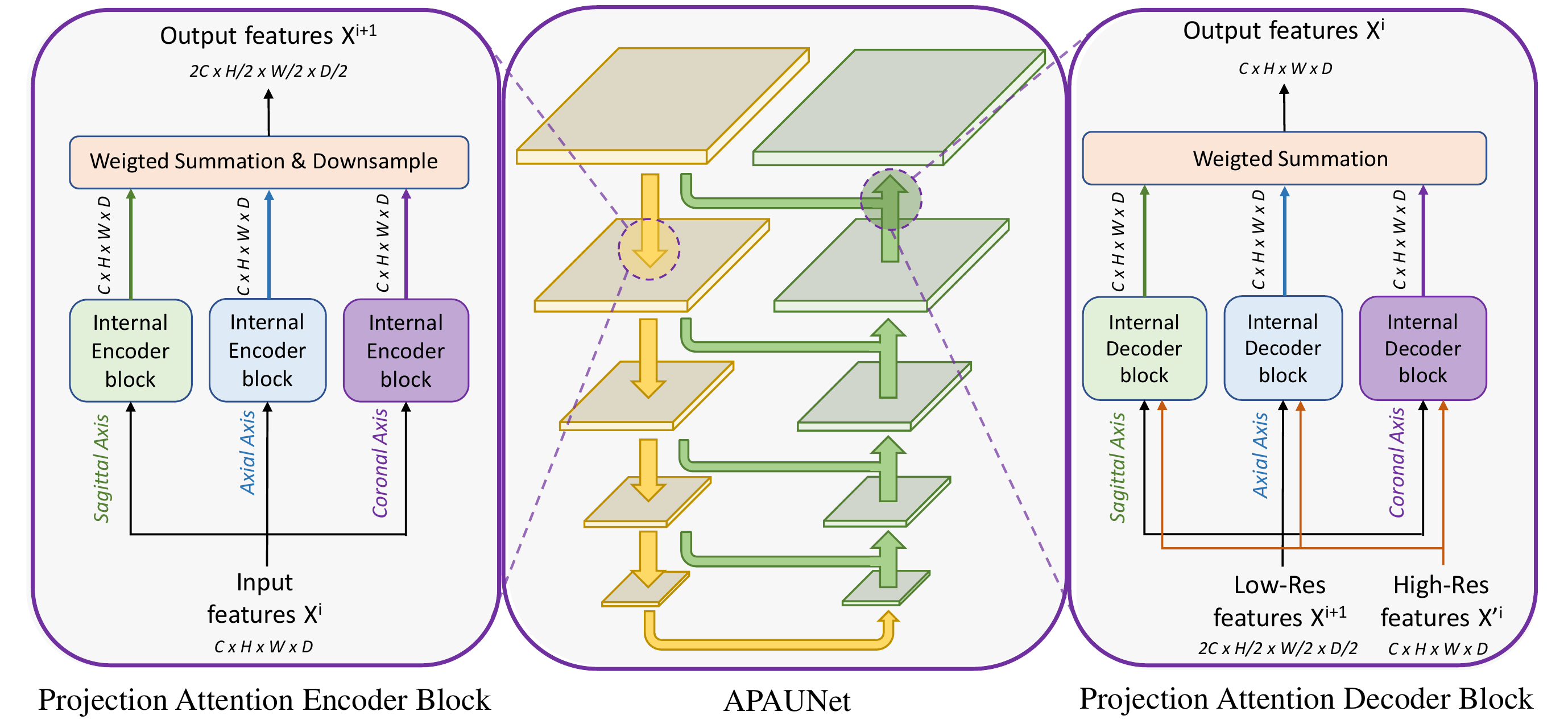}
    \caption{The overview of APAUNet. Taking 3D scans as the input, our APAUNet exploits the 3D-UNet as the backbone. 
    The yellow and green arrows denote the axis projection attention (APA) encoder and decoder blocks, respectively. 
    For the left part, within one APA encoder block, we use three parallel internal encoder (IE) to construct the 2D attention on each projected orthogonal planes to filter the redundant background features and then adaptively fuse the features from the three perspectives. 
    Correspondingly, for the right part, each APA decoder block merges the multi-scale features progressively to generate segmentation results.
	}
	\label{arch}
\end{figure}

Fig.~\ref{arch} illustrates the overall architecture of our APAUNet. 
Following the idea of 3D-UNet, our APAUNet consists of several axis projection attention (APA) encoder/decoder blocks with five resolution steps.
In this section, we first introduce the macro design of the APA encoder/decoder in Sec.~\ref{sec:PAencoder}, and then dig into the detailed block design in Sec.~\ref{sec:internal}.

\subsection{Axis Projection Attention (APA) Encoder and Decoder}
\label{sec:PAencoder}

The axis projection attention (APA) Encoder aims to extract contextual information of multiple resolution levels from different perspectives. The structure of APA Encoder is depicted in Fig.~\ref{arch} (left part).

In practice, given a 3D medical image $\mathit{I}\in\mathbb{R}^{C \times H \times W \times D}$, the APA Encoder extracts multi-level features at different resolution scales $\mathit{X^i} \in \mathbb{R}^{C_i \times \frac{H}{2^i} \times \frac{W}{2^i} \times\frac{D}{2^i}} $. 
For the $i$-th level, the input feature $ X^i $ is fed to three \textit{Internal Encoder (IE) blocks} in parallel to capture the contextual attention from three different perspectives. 
In order to capture the features of small targets more effectively, the original 3D features will be projected to three orthogonal 2D planes to extract the 2D spatial attention. 
And then, the learned 2D attentions and 3D feature maps are aggregated to enhance the feature representation. 
After obtaining fused features of the three axes, we use three learnable parameters $\beta_{i, a}$ to obtain the weighted summation of the enhanced features:
\begin{align}
    Y^{i} = \sum_{a=1}^{3}{\beta_{i, a} \cdot \text{IE}(X^i, a)}, \sum_{a=1}^{3}{\beta_{i, a}} = 1
\end{align}
where $\text{IE}(\cdot)$ is the operation of IE block and $a$ denotes the three orthogonal axes. 
This aggregation function can further help the network adaptively learn the importance of different projection directions to achieve asymmetric information extraction. 
Afterwards, another $1\times1\times1$ convolution and $2\times2\times2$ average pooling are applied to perform the down-sampling operation to get $X^{i+1}$, the input features of next level.

Similarly, the APA Decoder modules are used to extract and fuse multi-resolution features to generate segmentation results. The detailed design is shown in Fig.~\ref{arch} (right part). The APA Decoder block has a similar structure to the APA Encoder block, but takes two features with different resolutions as the input.

To be more specific, features of both low and high resolutions are fed into the APA Decoder block simultaneously, where the low resolution feature maps $X^{i+1} \in \mathbb{R}^{2C \times \frac{H}{2} \times \frac{W}{2} \times \frac{D}{2}}$ from the APA Decoder at the $(i+1)$-th level and the high resolution feature maps $ X^{'i} \in \mathbb{R}^{C \times H \times W \times D}$ from the APA Encoder at the $i$-th level.
Afterwards, the \textit{Internal Decoder (ID) block} aggregates both high and low resolution features to generate 2D contextual attentions from the three perspectives, then the 3D feature maps are fused with the 2D attentions to obtain 3D contextualized features, similar to that in the APA Encoder.  
To this end, the small-scale foreground information is better preserved, avoiding losing the crucial features. Finally, we take the weighted summation of three 3D contextualized features as the output features to next level.

\begin{figure}[t]
	\centering
	\includegraphics[width=0.6\linewidth]{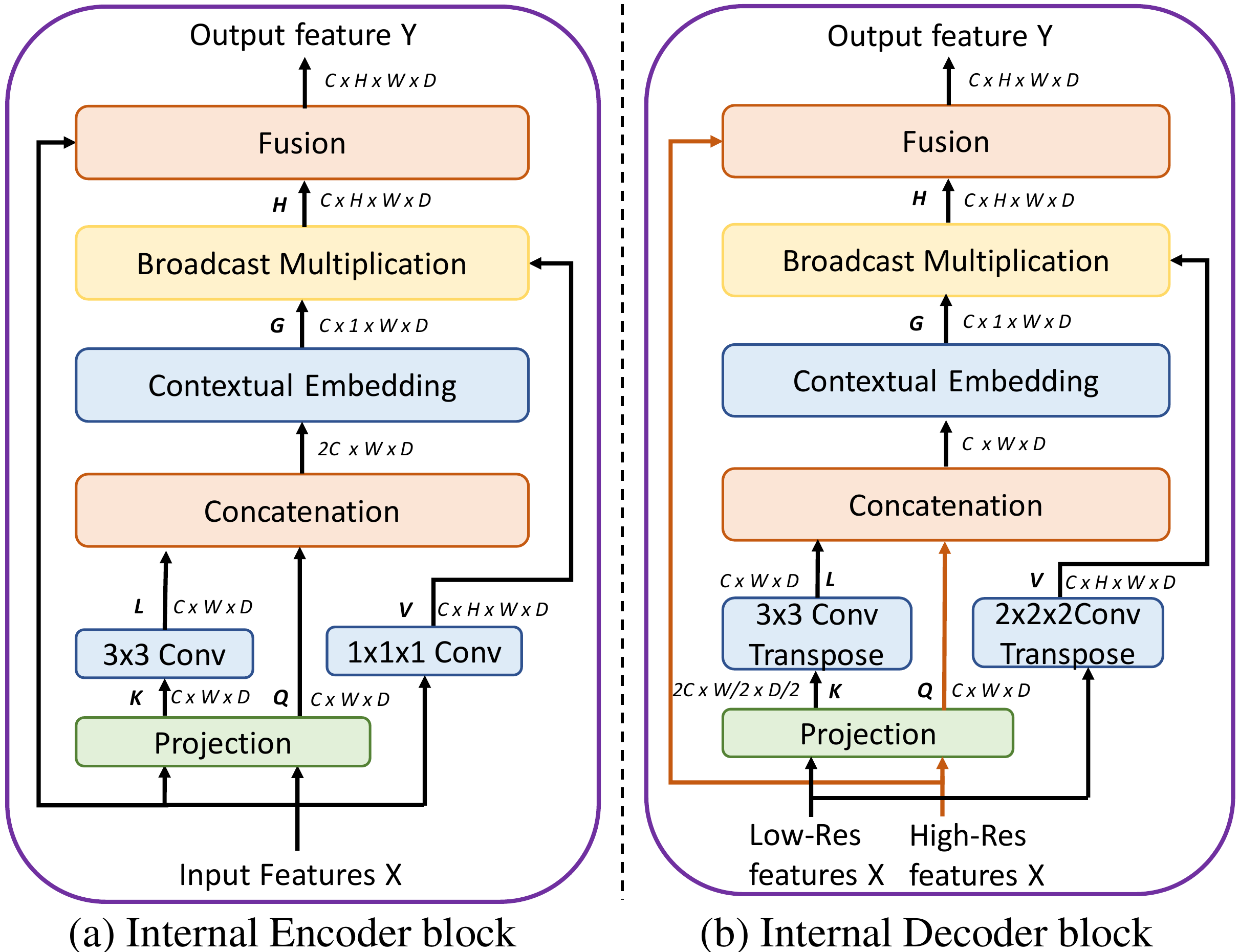}
	\caption{The detailed structure of Internal Encoder/Decoder (IE/ID) block used in the APA Encoder/decoder along the \textit{sagittal} ($H$) view respectively. 
	(a) The IE block. (b) The ID block. 
	They share a similar structure design while the ID block takes two inputs from different resolution levels.}
	\label{block}
\end{figure}

\subsection{Internal Encoder and Decoder Blocks}
\label{sec:internal}

Inspired by the Contextual Transformer (CoT)~\cite{li2021contextual}, we adopt 3 $\times$ 3 convolutions in our internal blocks to mine the 2D context information and follow the design of the attention mechanism in CoT blocks. 
Meanwhile, we introduce several refinement strategies to tackle the challenges of 3D medical images, especially for small lesions. 
The detailed structure of internal blocks are illustrated in Fig.~\ref{block}.

\subsubsection{Orthogonal Projection Strategy.} 

In the IE block, to better filter the irrelevant background and amplify the critical information of small lesions, we first project the input 3D features to three 2D planes. 
In particular, the 3D input feature $X \in \mathbb{R}^{H \times W \times D \times C}$ is projected to the \textit{sagittal}, \textit{axial} and \textit{coronal} planes of the Cartesian Coordinate System to generate keys ($K$) and queries ($Q$), whereas the values ($V$) keep the 3D shape. 
Taking the \textit{sagittal} view as an example, the input $X$ is projected to 2D to get keys and queries: $K, Q \in \mathbb{R}^{C \times W \times D}$, while $V \in \mathbb{R}^{C \times H \times W \times D}$ is obtained by a single $1\times1\times1$ convolution. 
Here we adopt the summation of global average pooling (GAP) and global max pooling (GMP) along the desired axis ($H$ in this case) as the projection operator:
\begin{align}
    K = Q &= {{\text{GAP}}}_{H}(X) + {{\text{GMP}}}_{H}(X)
\end{align}
\noindent The projected $K, Q$ are then used to learn different 2D contextual attentions along three orthogonal dimensions to better capture the key information of small lesions.

\subsubsection{Dimension Hybridization Strategy.} 

After the orthogonal projection, a $ 3 \times 3 $ group convolution with a group size of 4 is employed on $K$ to extract the local attention $L \in \mathbb{R}^{C \times W \times D}$, which contains the local spatial representation related to the neighboring keys. 
After that, we concatenate the local attention $L$ with $Q$ to further obtain the attention matrix $G \in \mathbb{R}^{C \times 1 \times W \times D}$ by two consecutive 1$\times$1 2D convolutions and dimension expend: 
\begin{align}
    L &= h(\text{GConv}_{3 \times 3}(K)) \\
	G &= \text{Unsqueeze}({\text{Conv}}_{1\times1}(h({\text{Conv}}_{1\times1}([L, Q]))))
\end{align}
\noindent where $\text{GConv}$ is the group convolution, $h(\cdot)$ denotes the normalization activation function, $[\cdot, \cdot]$ denotes the concatenation operation, and $\text{Unsqueeze}(\cdot)$ is the dimension expend operation. 
The attention matrix $G$ encodes not only the contextual information within isolated query-key pairs but also the attention inside the keys. 

Next, based on the 2D global attention $G$, we calculate the hybrid attention map $H \in \mathbb{R}^{H \times W \times D \times C} = V \odot G$, 
where $\odot$ denotes the broadcast multiplication, rather than the local matrix multiplication operation used in the original CoT blocks. 
This is because we empirically discovered that the local matrix multiplication operation does not contribute to any performance improvement under our 2D-3D hybridization strategy and consumes more GPU memory during training. 
Finally, the obtained hybrid attention maps are fused with the input feature $X$ by a selective attention~\cite{sknet} to get the output features $Y$. 
The complete structure of the IE block is illustrated in Fig.~\ref{block}(a). 

\subsubsection{Multi-Resolution Fusion Decoding.} 

In the ID block, in order to better obtain the multi-scale contextual information from the multi-resolution features, we integrate the upsampling operation into the attention extraction process. 
The design of ID block is similar to the IE block introduced above but takes two inputs from different resolution levels, 
where the high resolution features $X' \in \mathbb{R}^{C \times H \times W \times D}$ from the encoder are taken to produce the queries and the low resolution features from the previous decoder $X \in \mathbb{R}^{2C \times \frac{H}{2}\times\frac{W}{2}\times\frac{D}{2}}$ are taken to generate the keys and upsampled values. 
Then a $3 \times 3 \times 3$ transpose convolution is applied to upsample the keys to obtain the local attention $L$. 
Thus the decoder could fully capture the hybrid attention from various scales. 
The subsequent contextual extraction operations are similar to the IE block. 
And finally, we fuse the hybrid attention maps with the high resolution features $X'$ to generate the output features $Y$.
The detailed design of the ID block is illustrated in Fig.~\ref{block}(b).

\subsection{Loss Function}
\label{sec:loss}

Following the previous methods~\cite{cciccek20163d,xie2021segmenting}, 
we jointly use the Dice loss and the Cross Entropy loss to optimize our network. 
Specifically, 

\begin{align}
    \mathcal{L}_{total}&=\mathcal{L}_{dice}+\mathcal{L}_{CE} \\
    \mathcal{L}_{dice}&=1-\dfrac{2}{C}\sum_{j=1}^{C}\dfrac{\sum_{i=1}^{N}X_{ij}Y_{ij}}{\sum_{i=1}^{N}X_{ij}+\sum_{i=1}^{N}Y_{ij}} \\
    \mathcal{L}_{CE}&=-\frac{1}{N}\sum_{i=1}^N\sum_{j=1}^C(Y_{ij}\log X_{ij})
\end{align}

\noindent where $N$ denotes the total number of voxels, $C$ is the number of target classes, while $X$ and $Y$ are the softmax output of the prediction and the one-hot ground truth, respectively.

\section{Experiments}

\subsection{Datasets and Implementation Details}

Our APAUNet is evaluated on two 3D segmentation datasets: Synapse multi-organ segmentation dataset (BTCV)~\cite{miccaibtcv} and Medical Segmentation Decathlon (MSD) challenge~\cite{simpson2019large}. 
The synapse multi-organ segmentation dataset includes 30 cases of CT scans. Following the settings of \cite{missformer,TransUNet2021}, we split 18 cases for model training while the rest 12 cases are used as test set. 
We report the 95\% Hausdorff Distance (HD95) and Dice score (DSC) on 8 organs (aorta, gallbladder, spleen, left kidney, right kidney, liver, pancreas and stomach) following \cite{missformer,TransUNet2021}. 
While for MSD challenge we choose task 3 (Liver Tumours CT segmentation) and task 7 (Pancreas Tumour CT segmentation), since they contains more small targets. 
The Liver dataset contains 131 3D CT volumes with ground truth labels of both small liver tumours and large liver regions. 
While for the Pancreas dataset, it contains 282 3D CT volumes with labels of medium pancreas and small pancreas tumour regions. 
Since the MSD online test server is temporarily closed for submissions recently, we use the 5-fold cross-validation to evaluate our APAUNet on MSD datasets. 
Besides, we also reproduce the the comparison methods on MSD with the same settings if the results on these datasets are not reported/evaluated in their original papers. 

We implement our APAUNet on PyTorch and MONAI. 
For data preprocessing on MSD, we employ the method introduced in nnU-Net~\cite{isensee2018nnunet}, which crops the zero values and re-samples all cases to the median voxel spacing of their respective dataset. 
Besides, the third-order spline interpolation is used for image data and the nearest-neighbor interpolation for the corresponding segmentation mask.
During the training stage, we use a batch size of 2 and an input patch size of (96 $ \times $ 96 $ \times $ 96), where the patches are sampled from the original volumes. 
Specifically, to address the problem that the foreground targets are hard to be randomly cropped, we adopt an over-sampling strategy as introduced in~\cite{isensee2018nnunet} to enforce that more than half of the samples in a batch contain the foreground targets. 
All the patches are randomly flipped as data augmentation.
We use the SGD with an initial learning rate of 0.1, a momentum of 0.9, and a weight decay of 4e-5 with cosine annealing. 
The models are trained for 500 epochs in total with a 50-epochs warm-up. 
All the models are trained on an NVIDIA V100 GPU.

\setlength{\tabcolsep}{2pt}
\begin{table}[t]
    \centering
    \caption{
    Segmentation results on the synapse multi-organ segmentation dataset.
    Note: Aor: Aotra, Gall: Gallbladder, LKid: Kidney (Left), RKid: Kidney (Right), Liv: Liver, Pan: Pancreas, Spl: Spleen, Sto: Stomach.
    All the results are obtained from the corresponding papers, except nnUNet.
    }
    \begin{tabular}{c||cc||cccccccc}
        \hline
        \multirow{2}{*}{Method} & \multicolumn{2}{c||}{Average} & \multirow{2}{*}{Aor} & \multirow{2}{*}{Gall} & \multirow{2}{*}{LKid} & \multirow{2}{*}{RKid} & \multirow{2}{*}{Liv} & \multirow{2}{*}{Pan} & \multirow{2}{*}{Spl} & \multirow{2}{*}{Sto} \\
        \cline{2-3}
        & HD95 & DSC & & & & & & & & \\
        \hline \hline
        TransUNet & 32.62 & 77.49 & 87.23 & 63.13 & 81.87 & 77.02 & 94.08 & 55.86 & 85.08 & 75.62 \\
        CoTr & 27.38 & 78.08 & 85.87 & 61.38 & 84.83 & 79.36 & 94.28 & 57.65 & 87.74 & 73.55 \\
        UNETR & 23.87 & 79.57 & 89.99 & 60.56 & 85.66 & 84.80 & 94.46 & 59.25 & 87.81 & 73.99 \\
        MISSFormer & 19.20 & 81.96 & 86.99 & 68.65 & 85.21 & 82.00 & 94.41 & 65.67 & \textbf{91.92} & 80.81 \\
        nnFormer & 15.80 & 86.56 & 92.13 & 70.54 & 86.50 & 86.21 & 96.88 & 83.32 & 90.10 & \textbf{86.83} \\
        nnUNet & 13.69 & 86.79 & \textbf{93.20} & 71.50 & 84.39 & \textbf{88.36} & \textbf{97.31} & 82.89 & 91.22 & 85.47 \\
        \hline \hline
        APAUNet & \textbf{11.26} & \textbf{87.84} & 92.88 & \textbf{75.26} & \textbf{88.47} & 87.80 & 95.33 & \textbf{85.47} & 90.88 & 86.59 \\
        \hline
    \end{tabular}
    \label{res-synapse}
\end{table}

\setlength{\tabcolsep}{1pt}
\begin{table}[t]
    \centering
    \caption{
    Segmentation results (Dice score) on the MSD Liver and Pancreas Tumour datasets. 
    The results of nnU-Net and C2FNAS are from the MSD leaderboard, while other methods are reproduced/implemented by ourselves on fold-1, including our APAUNet. (* denotes the average score with the standard deviations of 5-fold cross-validation.)
    Latency (s) is measured by a sliding window inference on GPU.}
    \begin{tabular}{c||ccc||ccc||c}
        \hline
        \multirow{2}{*}{Method} & \multicolumn{3}{c||}{Liver} & \multicolumn{3}{c||}{Pancreas} & \multirow{2}{*}{Latency} \\ 
        \cline{2-4} \cline{5-7}
        & Organ & Cancer & Average & Organ & Cancer & Average & \\
        \hline \hline
        UNet3+  & 87.32 & 70.10 & 78.71 & 75.30 & 46.00 & 60.65 & 39.65 \\
        HFA-Net & 92.65 & 65.83 & 79.24 & 80.10 & 45.47 & 62.78 & 24.80 \\
        UTNet & 92.51 & 69.79 & 81.15 & 80.30 & 49.68 & 64.99 & 18.16 \\
		UNETR & 90.52 & 66.47  & 78.50 & 80.47 & 51.30  & 65.88 & 13.46 \\
		CoTr & 90.20 & 69.88 & 80.04 & 77.9 & 50.20 & 64.05 & 21.77 \\
		\hline \hline
		nnU-Net & 95.24 & \textbf{73.71} & \textbf{84.48} & 79.53 & 52.27 & 65.90 & - \\
		C2FNAS & 94.91 & 71.63 & 83.27 & 80.59 & 52.87 & 66.73 & - \\
		\hline \hline
		APAUNet & \textbf{96.10} & 72.50 & 84.30 & \textbf{83.05} & \textbf{55.21} & \textbf{69.13} & \textbf{13.36} \\ 
		APAUNet* & 95.56 {\tiny $\pm$0.35} & 71.99 {\tiny $\pm$0.81} & 83.78 {\tiny $\pm$0.35} & 82.29 {\tiny $\pm$1.73} & 54.93 {\tiny $\pm$1.64} & 68.61 {\tiny $\pm$1.63} & - \\
		\hline
    \end{tabular}
    \label{res-segmentation}
\end{table}

\subsection{Comparisons with Other Methods}

We first evaluate our APAUNet against several state-of-the-art methods, including TransUNet~\cite{TransUNet2021}, CoTr~\cite{CoTr2021}, UNETR~\cite{hatamizadeh2022unetr}, MISSFormer~\cite{missformer}, and nnFormer~\cite{zhou2021nnformer}. 
The results are shown in Tab.~\ref{res-synapse}. 
Our APAUNet achieves the best average dice score of 87.84 and the best average HD95 score of 11.26, 
and outperforms the second best method nnFormer by 4.54 and 1.28 on these two metrics respectively. 
For the class-wise results, our APAUNet also achieves the best dice scores in five of the categories and surpasses the second best method by 0.75 $\sim$ 4.72, while is only 1.55, 1.04 and 0.24 lower on Liver, Spleen and Stomach.

While on MSD datasets, we compare our APAUNet with HFA-Net~\cite{zheng2019hfa}, UTNet~\cite{gao2021utnet}, UNETR~\cite{hatamizadeh2022unetr}, UNet3+~\cite{9053405}, CoTr~\cite{CoTr2021}, C2FNAS~\cite{yu2020c2fnas} and nnUNet~\cite{isensee2018nnunet}. 
Since the MSD online leaderboard submission is temporarily closed at the time of the experiments, 
we focus on comparing our algorithm with the first 5 methods which are reproduced under the same settings, 
and also compare the 5-fold results with C2FNAS and nnUNet for reference.
Tab.~\ref{res-segmentation} shows the overall results of these methods. 
On Liver Tumour dataset, our APAUNet achieves a dice score of 96.10 on organ segmentation and 72.50 on cancer segmentation, 
which outperforms the second best method by 3.45 (HFA-Net) and 2.40 (UNet3+) on the dice score respectively. 
Also, the average dice score of our APAUNet is 84.30, which is 3.15 higher than UTNet. 
Similarly, the results of our APAUNet consistently outperform those of other methods on Pancreas dataset. 
Comparing with C2FNAS and nnUNet, our APAUNet surpasses both methods on all scores except Liver-cancer and Liver-average, where nnUNet uses additional test time augmentation, model ensembling and post-processing for better results. 
The results demonstrate the effectiveness of our APAUNet. 
More detailed results of 5-fold cross validation are shown in appendix.

\subsection{Results on Small Cases} 

\begin{figure}[t]
	\centering
	\includegraphics[width=\linewidth]{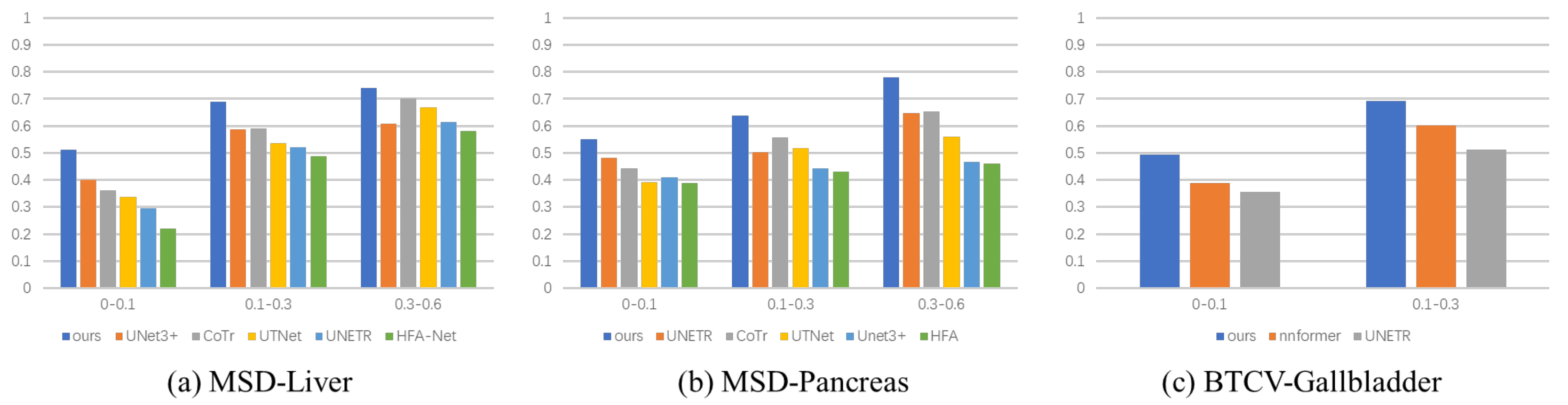}
	\caption{
	Statistical results of small cases on MSD and BTCV datasets. 
	Note that since most targets are concentrated in the range of 0-0.3\%, 
	in order to show the results more clearly, we divide 0-0.3\% into two intervals of 0-0.1\% and 0.1-0.3\%. 
	And also, the split validation set of BTCV does not contain any targets with size of 0.3-0.6\%. 
	}
	\label{res-smallcase}
\end{figure}

To verify the effectiveness of our APAUNet on small cases, we count the results of small targets with size less than 0.6\% on Liver, Pancreas and BTCV. 
Fig.~\ref{res-smallcase} shows the performance of our APAUNet and comparison methods. 
We observe that our APAUNet surpasses the second best results by 3.9 $ \sim $ 12.6 in absolute dice scores. 
Concretely, on the interval of 0 $\sim$ 0.1\%, our APAUNet achieves dice scores of 51.3, 55.1 and 49.4 on three datasets respectively, 
and has an improvement of 27.9\%, 14.3\% and 27.1\% compared to the second best methods. 
Similarly on the interval of 0.1 $\sim$ 0.3\%, the dice scores of our APAUNet are 16.6\%, 14.9\% and 15.2\% higher than the second best methods respectively. 
These exciting results further highlights the superiority of our APAUNet, especially on small target segmentation. 

\subsection{Visualization Results} 

\begin{figure}[t]
	\centering
	\includegraphics[width=0.75\linewidth]{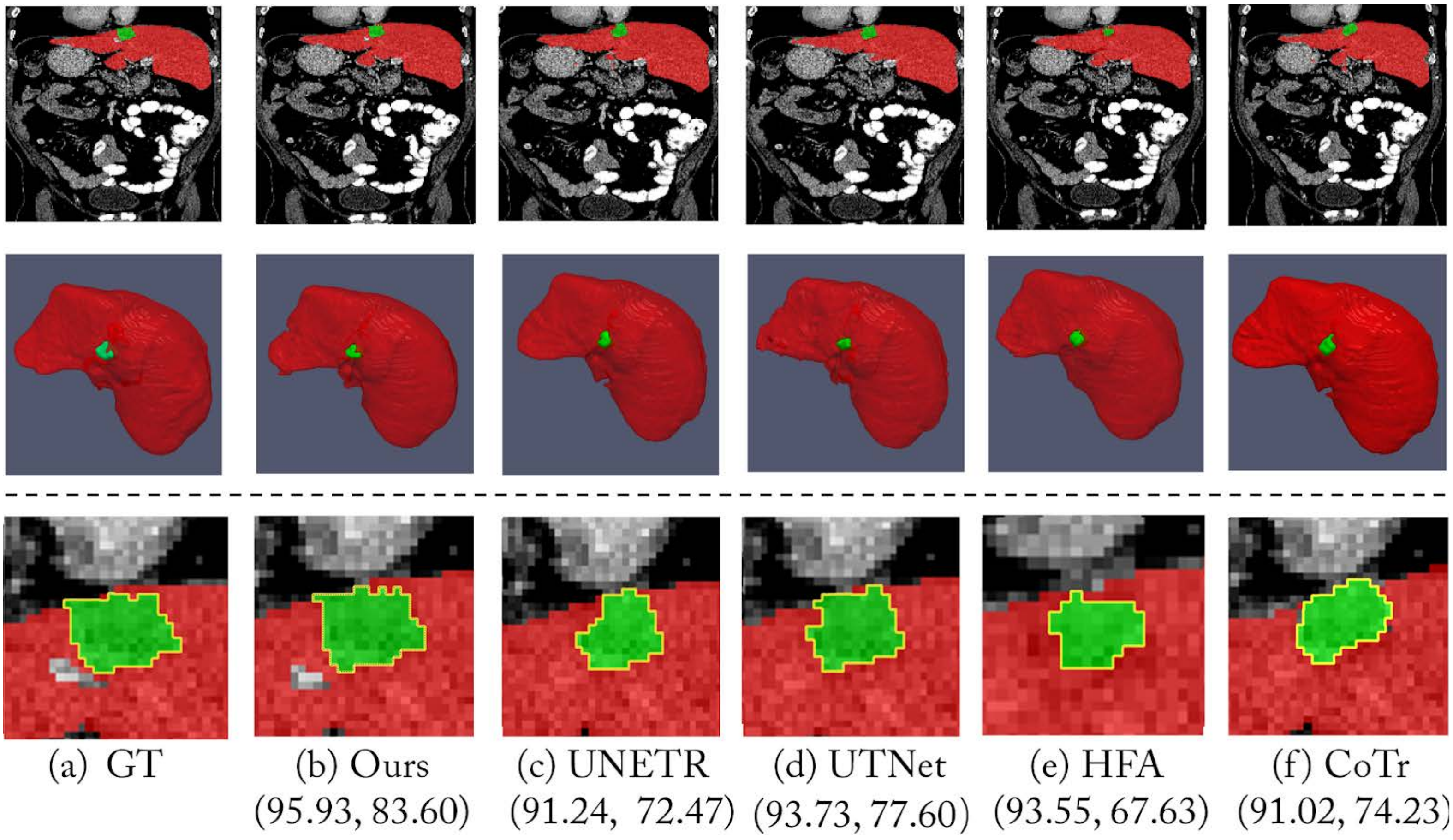}
	\caption{Visualization of our proposed APAUNet and other methods on Liver. 
	The upper two rows are the overall of segmentation results and the view of foreground targets, while the last row is an enlarged view of the segmentation results of small lesions.
	The dice scores $(organ, tumour)$ are given under each method.}
	\label{vis}
\end{figure}

\begin{figure}[t]
	\centering
	\includegraphics[width=0.75\linewidth]{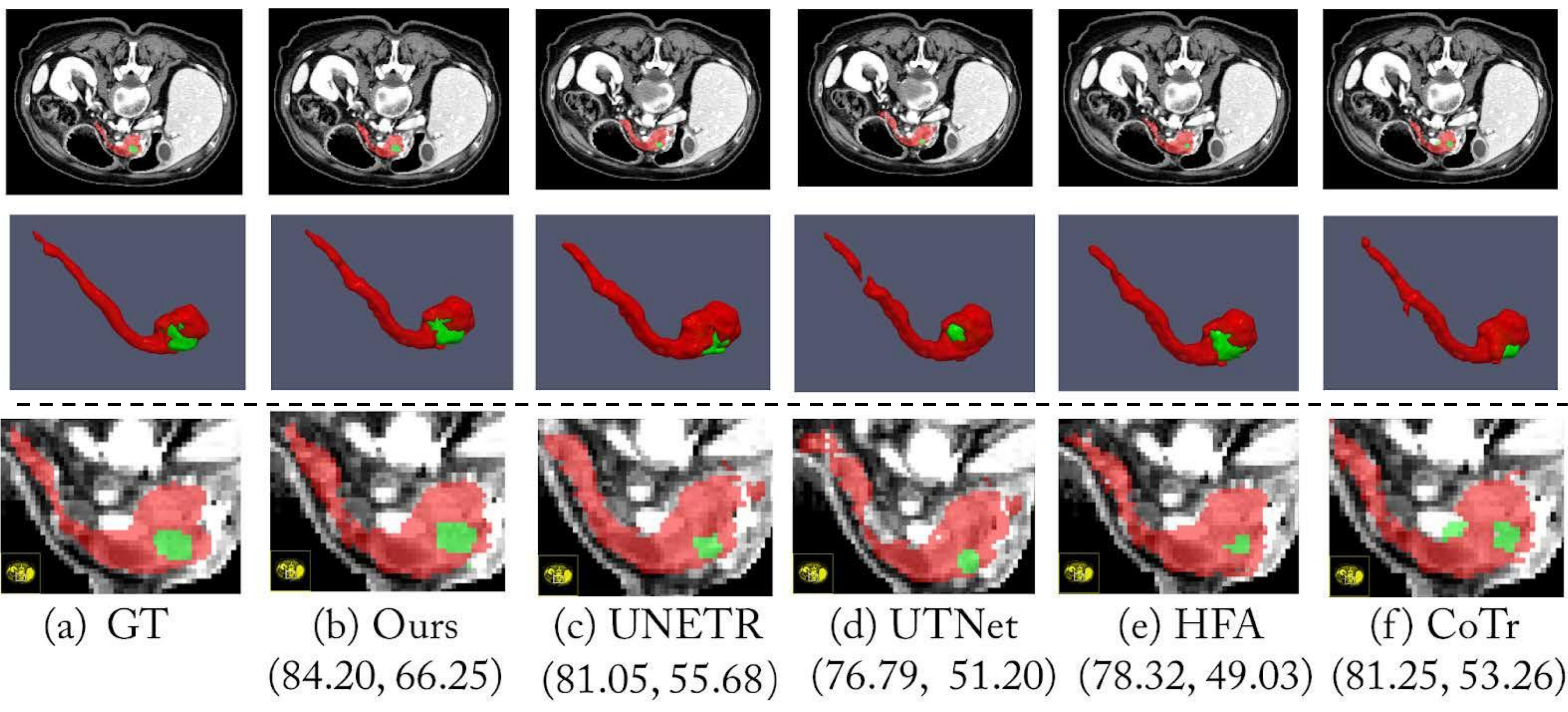}
	\caption{The visualization results of our APAUNet and comparison methods on Pancreas.}
	\label{more-vis-pan}
\end{figure}

In this part, we visualize some segmentation results of our APAUNet and compare with UNETR, UTNet, CoTr and HFA-Net on the Liver Tumour dataset in Fig.~\ref{vis} and on the Pancreas Tumour dataset in Fig.~\ref{more-vis-pan}. 
It can be observed that all the methods have good performance on large organ segmentation with a dice score of over 90 on liver and over 75 on pancreas. 
While on small lesions, UNETR, UTNet, CoTr and HFA-Net could only locate the position and are still unsatisfactory in the segmentation of both the boundary and the size. 
In comparison, our APAUNet gets clearer and more accurate boundaries and sizes, further verifying the substantial advantage of APAUNet in medical image segmentation, especially for small targets.

\subsection{Analysis and Ablation Studies}

In this section, a variety of ablation studies are performed to thoroughly verify the design our APAUNet and validate the performance under different settings.
More ablation results can be found in appendix.

\subsubsection{2D-3D Hybridization Strategy} 

To verify the effectiveness of our 2D-3D hybridization strategy, we conduct several ablation experiments. 
First, we replace the 2D operators in the original CoT block with 3D operators, which is a pure 3D version of CoTNet (CoT-3D). 
For the pure 2D version (CoT-2D), we project all the $K, Q$ and $V$ to 2D planes. 
The results are shown in Tab.~\ref{ab-hybrid}. Compared with CoT-3D, our APAUNet achieves an improvement of 7.13 and 8.29 on average dice scores. 
Different from natural images, 3D medical CT scans contain more excessive background information that hinders the learning of contextual attention, 
while our 2D-3D hybridization strategy can filter the redundant information to enhance the performance. 
Also, APAUNet outperforms CoT-2D by a large margin, which indicates that it is suboptimal for the model to utilize only 2D attentions for 3D segmentation tasks. 

\setlength{\tabcolsep}{4pt}
\begin{table}[t]
	\centering
    \caption{Ablation experiments on the 2D-3D hybridization strategy, projection operators and the weighted importance. 
    Avg, Max and Conv refer to the avgpooling, maxpooling and convolution operations, respectively.
    It can be seen that the effectiveness of the design of our APAUNet.}
	\begin{tabular}{c||ccc||ccc}
	    \hline
	    \multirow{2}{*}{Method} &\multicolumn{3}{c||}{Liver}&\multicolumn{3}{c}{Pancreas}\\
		\cline{2-4} \cline{5-7}
		& Organ & Cancer & Avg & Organ & Cancer & Avg\\ 
		\hline \hline
		CoT-2D & 82.66 & 59.08 & 70.87 & 74.30 & 43.28 & 58.79\\
		CoT-3D & 91.08 & 63.26 & 77.17 & 75.40 & 46.28 & 60.84\\
		\hline \hline
		Conv & 85.20 & 68.81 & 77.00 & 80.24 & 51.33 &  65.78 \\ 
		Max & 87.34 & 69.95 & 78.65 & 80.47 & 53.28 & 66.87 \\  
		Avg & 95.89 & 71.68 & 83.79 & 82.39 & \textbf{55.48} & 68.94\\ 
		\hline \hline
		Mean & 95.63 & 71.55 & 83.59 & 81.67 & 54.30 & 67.99 \\
		\hline \hline
		APAUNet & \textbf{96.10} & \textbf{72.50} & \textbf{84.30}  &\textbf{83.05} & \textbf{55.21} & \textbf{69.13} \\ \hline
	\end{tabular}
    \label{ab-hybrid}
\end{table}

\subsubsection{Projection Operations} 

For the choice of projection operators, we conduct several experiments to figure out the best choice in \{avgpooling, maxpooling, convolution\}. 
For the fair comparison of the channel-wise operation of pooling, here we use a depth-wise convolution to perform the projection operation. 
Experiment results are shown in Tab.~\ref{ab-hybrid}. 
Unexpectedly, the results of using convolution are the worst, despite its learnable parameters that make it more flexible than the pooling operators. 
We conjecture that the convolution lacks the ability to capture important information and is easily disturbed by the noise due to the high proportion of background information in medical images, even though with large kernel sizes. 
Fortunately, pooling operators could directly process the global information. 
Specifically, avgpooling involves all parts of the image to get the global statistics and maxpooling focuses only on the most salient part. 
By combining the complementarity of avgpooling and maxpooling, better performance can be achieved.
Hence, we adopt the summation of avgpooling and maxpooling as the projection operation in our APA Encoder and Decoder.

\subsubsection{Weighted Importance of Projection} 

Recall that there are three learnable weights during the fusion of features generated from different projection directions. 
Here, we conduct an experiment to study the effect of these weights. 
The results are shown in Tab.~\ref{ab-hybrid}. 
Compared to the average mean, our APAUNet could gain an extra improvement of 0.47 $\sim$ 1.38 on both datasets with the learnable weighted importance, which shows the feasibility of our asymmetric feature extraction on different projection axes. 
At the same time, this interesting phenomenon shows that the amount of information contained in different perspectives in the 3D structure is different, which motivates us to discover better 2D-3D fusion strategies.
More detailed experimental results can be found in the appendix.

\section{Conclusion}

In this paper, we propose a powerful network for the 3D medical segmentation task, called Axis Projection Attention Network (APAUNet). 
In order to deal with the highly imbalanced targets and background, we leverage an orthogonal projection strategy and dimension hybridization strategy to build our APAUNet. 
Extensive experiments on the BTCV and MSD datasets demonstrate the promising results on 3D medical segmentation tasks of our APAUNet, and the superior results on small targets show the effectiveness of our APA Encoder/Decoder blocks. 
In our experiments, we also show that the importance of different perspectives is different, which motivates us to look deeper into the projection strategy and dimension hybridization pattern. 
For the future work, we would like to enhance the performance of small target segmentation by exploring more advanced projection techniques and designing more efficient dimension fusion strategies.


\subsubsection{Acknowledgements} 

This work was supported in part 
by the National Key R\&D Program of China with grant No.2018YFB1800800, 
by the Basic Research Project No. HZQB-KCZYZ-2021067 of Hetao Shenzhen HK S\&T Cooperation Zone, 
by NSFC-Youth 61902335, 
by Shenzhen Outstanding Talents Training Fund, 
by Guangdong Research Project No.2017ZT07X152 and No.2019CX01X104, 
by the Guangdong Provincial Key Laboratory of Future Networks of Intelligence (Grant No.2022B1212010001), 
by zelixir biotechnology company Fund, 
by the Guangdong Provincial Key Laboratory of Big Data Computing, The Chinese University of Hong Kong, Shenzhen,
by Tencent Open Fund, and by ITSO at CUHKSZ.

\appendix

\section{Results of 5-fold Cross-validation}

In the experiment section, we evaluate our APAUNet using 5-fold cross validation on MSD-Liver and MSD-Pancreas, here we present the complete results in Tab.~\ref{exp-5fold}. 
The mean results of our APAUNet on the Liver dataset are 95.56, 71.99 and 83.78 respectively, and the standard deviations are 0.35, 0.81 and 0.35 respectively. 

Similarly, on the Pancreas dataset, our APAUNet achieves the mean results of 82.29, 54.93 and 68.61 respectively and the standard deviations are 1.73, 1.64 and 1.63 respectively. The standard deviations are slightly higher than those on the Liver dataset, since the target size of the Pancreas dataset is much smaller than that of the Liver dataset. 
Overall, the results of experiments demonstrate the effectiveness and stability of our APAUNet.

\begin{table}[h!]
	\centering
	\setlength\tabcolsep{4pt}
	\caption{The complete segmentation results of 5-fold validation on the Liver and Pancreas Tumour datasets.}
	\begin{tabular}{c|ccc|ccc}
		\hline
		{\multirow{2}{*}{Method}} & \multicolumn{3}{c|}{liver} & \multicolumn{3}{c}{Pancreas} \\ \cline{2-7} 
		{} &  {Organ} &  {Cancer} & Avg &  {Organ} &  {Cancer} & Avg \\ 
		\hline \hline
		fold-1 &  {96.10} &  {72.50}  & \textbf{84.30} &  {83.05} &  {55.21}  & 69.13 \\ \hline
		fold-2 &  {\textbf{96.23}} &  {71.45}  & 83.84 &  {80.16} &  {53.20}  & 66.68 \\ \hline
		fold-3 &  {94.87} &  {70.66}  & 82.77 &  {\textbf{83.40}} &  {\textbf{56.66}}  & \textbf{70.03} \\ \hline
		fold-4 &  {95.48} &  {72.66}  & 84.07 &  {81.88} &  {54.30}  & 68.09 \\ \hline
		fold-5 &  {95.12} &  {\textbf{72.69}}  & 83.91 &  {82.94} &  {55.27}  & 69.11 \\ 
		\hline \hline
		Mean &  {95.56} &  {71.99}  & 83.78 &  {82.29} &  {54.93}  & 68.61 \\ \hline
		STD &  {$\pm$0.35} &  {$\pm$0.81}  & $\pm$0.35 &  {$\pm$1.73} &  {$\pm$1.64}  & $\pm$1.63 \\ \hline
	\end{tabular}
	\label{exp-5fold}
\end{table}

\section{Detailed Computational Analysis}

Here we show the computational cost analysis in Tab.~\ref{compute}. 
The inference time is measured by a sliding window inference with patch size of 96 on GPU. 
Our APAUNet has a similar training GPU memory with HFA-Net/UNETR and a similar inference GPU with UTNet/CoTr, and achieves the lowest inference time. 

\setlength{\tabcolsep}{2pt}
\begin{table}[t]
	\centering
	\caption{Computational analysis of our APAUNet and other methods. 
		Parameters (M), training/inference GPU consumption (G) and inference time (s).}
	\begin{tabular}{c||cccc}
		\hline
		Method & Param.(M) & Training GPU(G) & Inference GPU(G) & Inference time(s) \\
		\hline \hline
		UNet3+ & 94.67 & 30.37 & 14.33 & 39.65 \\
		HFA-Net & 40.7 & 12.02 & 8.54 & 24.80 \\
		UTNet & \textbf{35.26} & 9.72 & 6.01 & 18.16 \\
		UNETR & 92.24 & 13.78 & \textbf{3.87} & 13.46 \\
		CoTr & 41.86 & \textbf{8.69} & 6.91 & 21.77 \\
		\hline \hline
		APAUNet & 76.94 & 11.74 & 6.42 & \textbf{13.36} \\
		\hline
	\end{tabular}
	\label{compute}
\end{table}

\section{Effect of Each Strategy}

In this section, we conduct several ablation studies by adding our strategies step by step. 
The results are shown in Tab.~\ref{ablation}.
It can be observed that all the strategies are necessary for the final performance. 
Combining all the proposed strategies, our model achieves the new state-of-the-art performance. 
In addition, our projection strategy has lower memory consumption despite it needs replicate the 3D fused features three times.

\setlength{\tabcolsep}{2pt}
\begin{table}[t]
	\centering
	\caption{Ablation experiments under different settings. OP - orthogonal projection, DH - dimension hybridization, MRF - multi-resolution fusion.}
	\begin{tabular}{c||ccc||c||ccc||ccc}
		\hline
		\multirow{2}{*}{Settings} & \multirow{2}{*}{OP} & \multirow{2}{*}{DH} & \multirow{2}{*}{MRF} & \multirow{2}{*}{GPU (G)} &\multicolumn{3}{c||}{Liver}&\multicolumn{3}{c}{Pancreas}\\
		\cline{6-8} \cline{9-11}
		& & & & & Organ & Cancer & Avg & Organ & Cancer & Avg\\ \hline \hline
		1 & \checkmark & $\times$ & $\times$ & 11.75 & 87.42 & 63.88 & 75.65 & 75.40 & 46.28 & 60.84 \\ 
		2 & $\times$ & $\times$ & \checkmark & 13.02 & 90.15 & 62.41 & 76.28 & 76.80 & 47.13 & 61.97 \\ 
		3 & \checkmark & \checkmark & $\times$ & 11.72 & 94.81 & 69.32 & 82.07 & 80.30 & 53.92 & 67.11 \\ 
		4 & \checkmark & $\times$ & \checkmark & 11.71 & 92.76 & 67.88 & 80.32 & 77.83 & 49.03 & 63.43 \\ 
		\hline \hline
		APAUNet & \checkmark & \checkmark & \checkmark & 11.74 & \textbf{96.10} & \textbf{72.50} & \textbf{84.30} & \textbf{83.05} & \textbf{55.21} & \textbf{69.13}\\ 
		\hline
	\end{tabular}
	\label{ablation}
\end{table}

\section{Weighted Importance of Projection}

As we mentioned in the main paper, adding learnable weights could lead to a performance boost. 
In this part, we present the detailed results of the learnable weights on the Liver and Pancreas datasets. 
Tab.~\ref{ab-weight} shows the weights learned after the training process on Liver and Pancreas dataset, respectively.
It can be observed from the results that the weights of the shallow encoders (1-3) in all directions are relatively balanced, while the encoders of deep level may selectively emphasize a certain projection axis, e.g., 0.63 of Encoder-4 on \textit{axial}-axis on the Liver dataset and 0.46 of Encoder-4 on \textit{sagittal}-axis on the Pancreas dataset.
We conjecture that the encoder path is mainly used for feature extraction and analysis, thus the importance weight will not be particularly inclined to a certain projection axis.

While on the decoder path, the phenomenons of importance selection are more obvious. 
Most decoders will tend to choose one or two of the projection dimensions, e.g., \textit{sagittal}-axis of Decoder-2 and \textit{axial}-axis of Decoder-3 on the Liver dataset, \textit{axial}-axis of Decoder-1 and Decoder-2 on the Pancreas dataset. 
We analyze that the decoder path is used to synthesize multi-scale features to obtain the segmentation prediction, thus it may highlight more salient features and filter the redundant information.

\begin{table}[ht]
	\centering
	\setlength\tabcolsep{4pt}
	\caption{The detailed results of learned importance weights.}
	\begin{tabular}{c||c||ccc}
		\hline
		Dataset & Level & Sagittal & Axial & Coronal \\ \hline \hline
		\multirow{9}{*}{Liver} & Encoder-1 & 0.28 & 0.37 & 0.35 \\ \cline{2-5} 
		& Encoder-2 & 0.31 & 0.34 & 0.35 \\ \cline{2-5} 
		& Encoder-3 & 0.31 & 0.30 & 0.38 \\ \cline{2-5} 
		& Encoder-4 & 0.27 & \textcolor{red}{0.63} & 0.10 \\ \cline{2-5}
		& Encoder-5 & 0.20 & 0.39 & 0.39 \\ \cline{2-5}
		& Decoder-1 & 0.36 & 0.30 & 0.34 \\ \cline{2-5}
		& Decoder-2 & \textcolor{red}{0.70} & 0.16 & 0.14 \\ \cline{2-5}
		& Decoder-3 & 0.09 & \textcolor{red}{0.78} & 0.13 \\ \cline{2-5}
		& Decoder-4 & 0.30 & 0.42 & 0.28 \\ 
		\hline \hline
		\multirow{9}{*}{Pancreas} & Encoder-1 & 0.33 & 0.34  & 0.33  \\ \cline{2-5} 
		& Encoder-2 & 0.22 & 0.38 & 0.40 \\ \cline{2-5} 
		& Encoder-3 & 0.30 & 0.37 & 0.33 \\ \cline{2-5} 
		& Encoder-4 & \textcolor{red}{0.46} & 0.24 & 0.30 \\ \cline{2-5}
		& Encoder-5 & 0.18 & 0.44 & 0.38 \\ \cline{2-5}
		& Decoder-1 & 0.17 & \textcolor{red}{0.52} & 0.31 \\ \cline{2-5}
		& Decoder-2 & 0.22 & \textcolor{red}{0.67} & 0.11 \\ \cline{2-5}
		& Decoder-3 & 0.31 & 0.18 & \textcolor{red}{0.51} \\ \cline{2-5}
		& Decoder-4 & 0.27 & 0.34 & 0.37 \\ \hline
	\end{tabular}
	\label{ab-weight}
\end{table}

\section{More Visualization Results}

\begin{figure}[ht]
	\centering
	\includegraphics[width=0.8\linewidth]{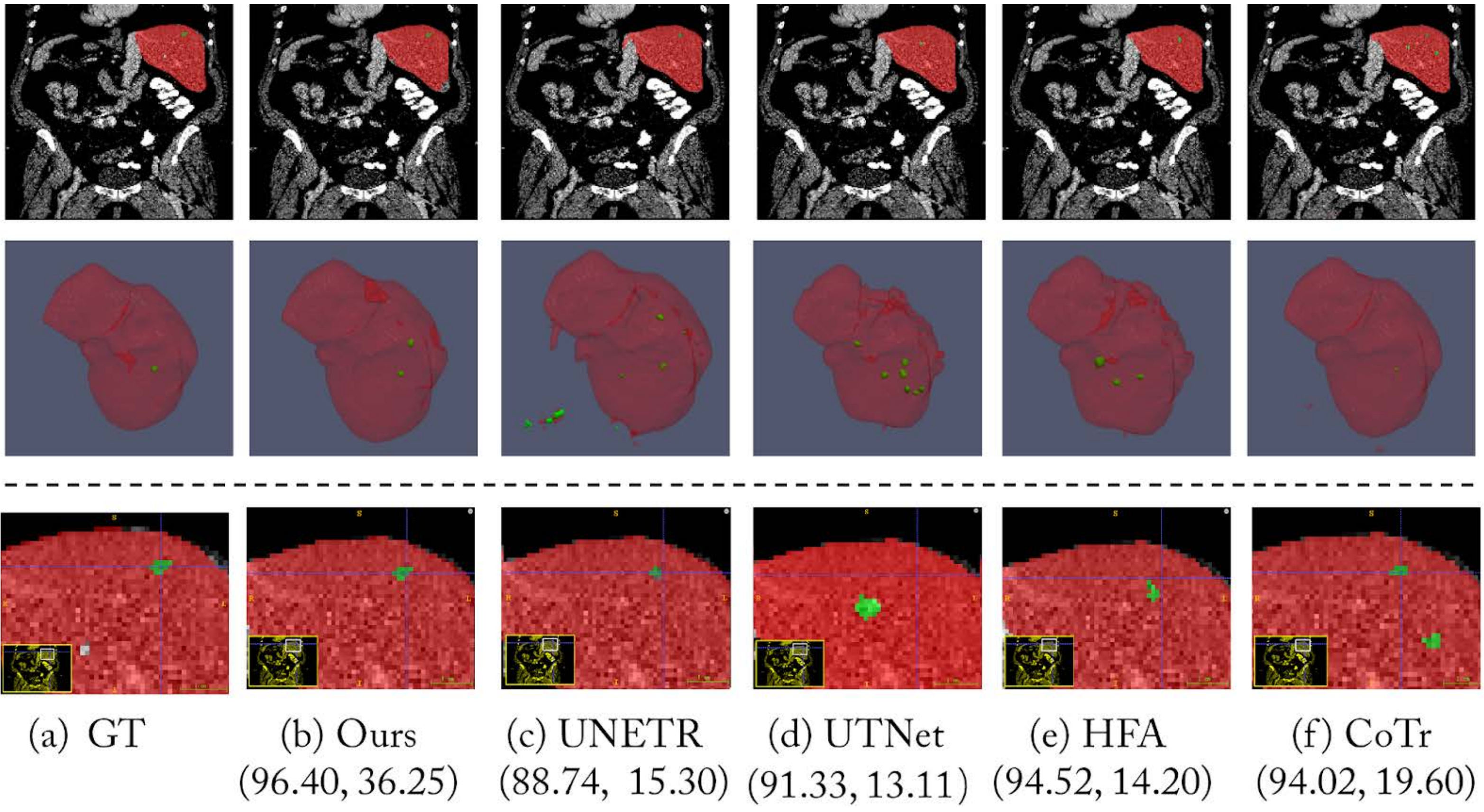}
	\caption{More visualization results of our APAUNet and comparison methods on Liver.}
	\label{more-vis-liver}
\end{figure}

\begin{figure}[ht]
	\centering
	\includegraphics[width=0.8\linewidth]{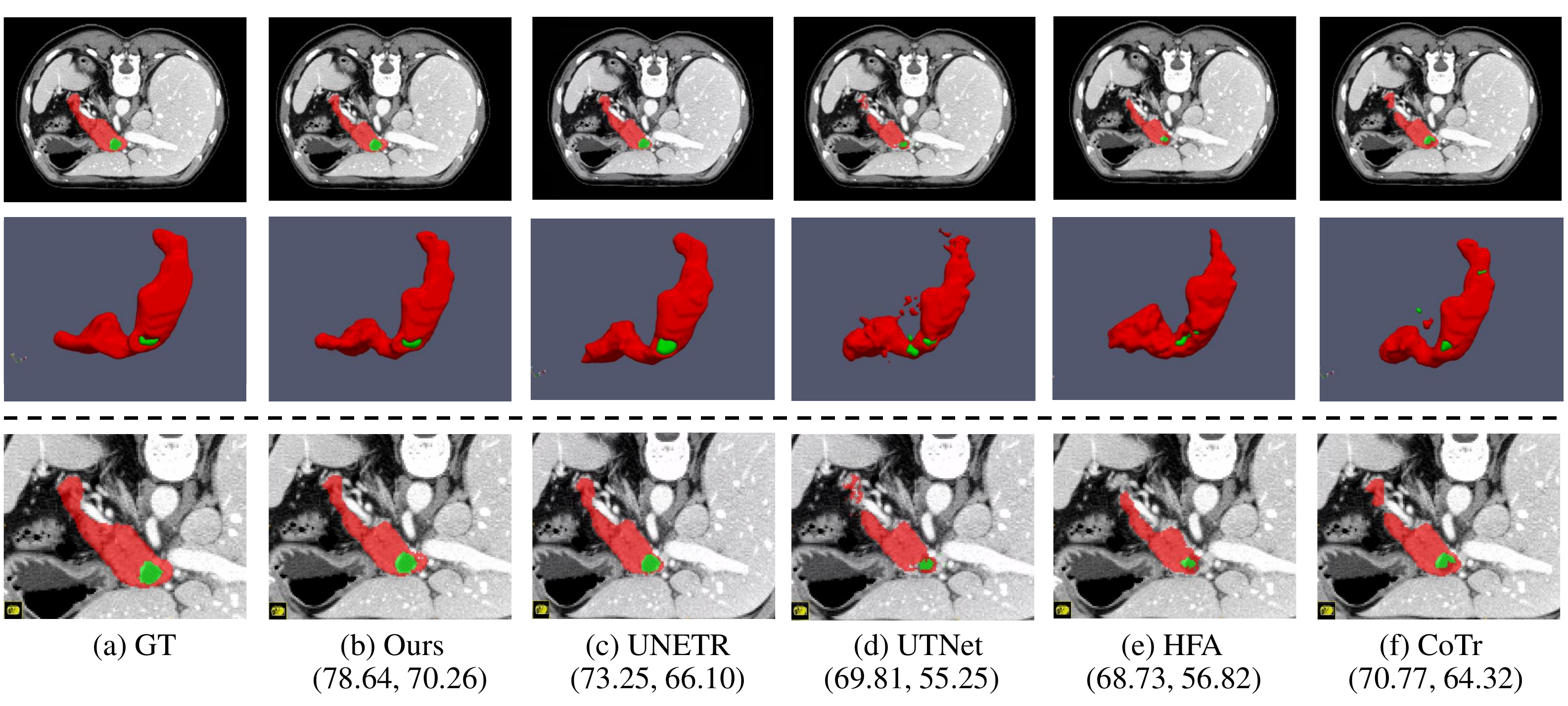}
	\caption{More visualization results of our APAUNet and comparison methods on Pancreas.}
	\label{more-vis-pancreas}
\end{figure}

\begin{figure}[ht]
	\centering
	\includegraphics[width=0.65\linewidth]{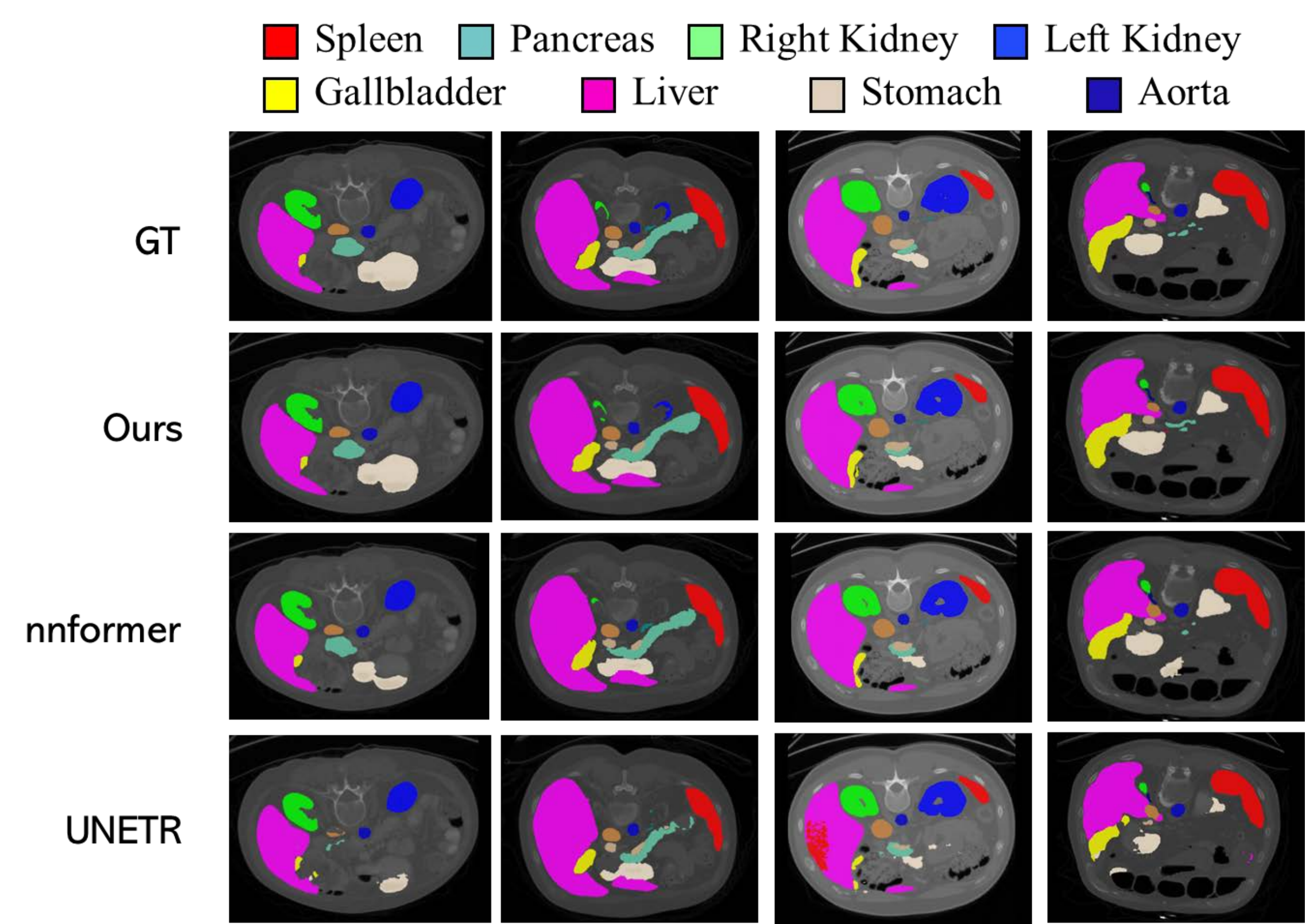}
	\caption{More visualization results of our APAUNet and comparison methods on BTCV.}
	\label{more-vis-btcv}
\end{figure}

In this part, we demonstrate more visualization results of our APAUNet on MSD-Liver, MSD-Pancreas and BTCV in Fig.~\ref{more-vis-liver}, Fig.~\ref{more-vis-pancreas} and Fig.~\ref{more-vis-btcv}. 

\end{document}